# How Prolific Sellers Self-Present:

# Dissecting the Communication Patterns of 1.6 Million Reverb Listings

David M. Markowitz[1]

**Affiliations**

[1] Department of Communication, Michigan State University, East Lansing, MI 48824

**Corresponding Author**

David M. Markowitz

404 Wilson Road

Michigan State University

East Lansing, MI 48824

dmm@msu.edu

## Abstract

The current paper draws on self-presentation theory and warranting theory to evaluate how the language patterns in an online marketplace reflect seller status (i.e., a prolific seller versus an everyday seller). Using 1.6 million musical instrument listings from Reverb.com in search of content, style, and structural differences in seller product descriptions, the evidence suggested prolific sellers tend to focus more on objective and functional aspects of a product (e.g., its features and specifications) and less on subjective characteristics like tone, relative to everyday sellers. Prolific sellers also communicated in a more narrative-like style, which was driven by an elevated use of personal pronouns, and they used longer descriptions than everyday sellers. Therefore, *what* prolific sellers focus on tends to be quite technical, but *how* they communicate this information is typical of a story that is told to potential buyers. Implications for self-presentation theory and warranting theory are discussed.

**How Prolific Sellers Self-Present:**

**Dissecting the Communication Patterns of 1.6 Million Reverb Listings**

Words are windows into the psychology of communicators. Decades of evidence in the social sciences, for example, have examined how language patterns connect to meaningful aspects of the human condition, including psychological well-being (Cohn et al., 2004; Edwards & Holtzman, 2017), personality traits (Ireland & Mehl, 2014; Kern et al., 2014), emotion (Sun et al., 2020; Vine et al., 2020), and many others (for extensive reviews, see Boyd & Markowitz, 2025; Boyd & Schwartz, 2021; Dehghani & Boyd, 2022; Pennebaker, 2011; Pennebaker et al., 2003; Tausczik & Pennebaker, 2010). Indeed, a keen interest in psychology of language research has been to develop social and psychological profiles of the individual mind through words. Scholars often use computational techniques to gather and analyze language data with the hope of trying to understand how words reveal psychological aspects of one's lived experience. Meaningful insights, which may have been difficult to obtain without text analysis, have been developed in how emotions can spread online (Coviello et al., 2014; Kramer et al., 2014), how collective action is formed (De Choudhury et al., 2016), how people learn and process information (Pennebaker et al., 2014), and what makes for lasting interpersonal relationships (Ireland et al., 2011). If there are words to analyze, there are pathways toward appreciating how people think, feel, and behave psychologically.

The current work seeks to build on this rich scholarly history to examine the psychology of prolific sellers on an active online marketplace, Reverb.com. This is a marketplace for musical instrument and musical gear sales, making it a fertile testing ground to understand how prolific sellers, defined as those who sell an above-average number of products, communicate about their products compared to everyday sellers. By evaluating the content, style, and structure of seller product descriptions, the current work aims to offer a lens into the psychology of those who sell a lot of products versus few products, and to connect such indicators to self-presentation theory (Goffman, 1959, 1967), warranting theory (Parks, 2011; Walther & Parks, 2002), and impression management processes that theoretically undergird them.

This work is timely and important for several reasons. First, online marketplaces have become dominant platforms of economic and social exchange, using text as often the primary channel through which people can evaluate one another (and the products being sold). On platforms like Reverb, a product description reflects details of what is being sold, and it is also a reflection of the seller (e.g., who they are, what they represent, what they are trying to communicate). Understanding the language of prolific sellers therefore has consequences that extend self-presentation processes to understand how top communicators manage impressions of themselves online through their words within product descriptions. Second, the contrast between prolific and everyday sellers offers a naturally-occurring comparison of expertise at scale, which is difficult to construct or manufacture in the laboratory. Identifying how high-volume sellers manage impressions (e.g., what they emphasize, how they structure their product descriptions) clarifies which self-presentational strategies may be associated with marketplace success and offers guidance for similar sellers and platforms. Together, this work is interested in advancing a deeper theoretical understanding of self-presentation, impression management, and warranting (e.g., why some claims are believed while others are discounted) through words, and this is achieved through a large-scale evaluation of music product descriptions from prolific (high volume) and everyday (low volume) sellers.

## A Primer on Psychology of Language Research: The Words-As-Attention Framework

The idea of using words to make psychological inferences from prolific versus everyday sellers is rooted in psychology of language research. This subfield argues words are markers of one's psychological focus and attention (Arnold, 1910; Boyd & Markowitz, 2025; Boyd & Schwartz, 2021; Pool, 1959). Put another way, a person who communicates with a high rate of positive emotion terms is *attending to* a positive affective state rather than necessarily *feeling* positive (see Markowitz, 2024). This words-as-attention framework to study the psychology of language has received substantial support and has been the subject of hundreds, if not thousands of studies (Boyd & Markowitz, 2025; Boyd & Schwartz, 2021; Dehghani & Boyd, 2022; Pennebaker, 2011; Pennebaker et al., 2003; Tausczik & Pennebaker, 2010).

What types of words do scholars evaluate when attempting to make psychological inferences from language data? There are three main classes of words that are often investigated. The first is content words, which typically include nouns and verbs. Content words describe *what* a person is communicating about and through various topic modeling techniques (e.g., Blei et al., 2003; Markowitz, 2021), content words can cluster to form coherent themes from texts. For example, a study by Blackburn et al. (2018) used a food-related subreddit to analyze what people wrote about in healthy versus unhealthy cities in the US. The evidence suggested people in healthy cities wrote more about home cooking and less about alcohol than people in unhealthy cities. Content patterns are essential for understanding what people are communicating, which is different from one's communication style and the second class of words often investigated.

Communication style is often assessed through function words (also known as style words; Chung & Pennebaker, 2007). Style words represent nearly half of all words that people communicate in English (Rochon et al., 2000), yet they are infrequent in terms of how many style words there are in totality (e.g., approximately, there are only 400 style words) (Baayen et al., 1995). They include words like articles (e.g., *a*, *the*), prepositions (e.g., *above*, *below*), and pronouns (e.g., *I*, *she*) to describe *how* a person is communicating as opposed to *what* a person is communicating about (content). A wealth of evidence suggests style words are deeply connected to a range of social and psychological dimensions (Pennebaker, 2011). For instance, recent scholarship suggests first-person singular pronouns (e.g., *I*, *me*, *my*) are positively associated with distress stemming from romantic breakups (Seraj et al., 2021) and negatively associated with academic performance in college and one's interest in deep thinking (Markowitz, 2023a; Pennebaker et al., 2014). It is important to draw attention to this academic performance example because it is foundational to the current empirical effort. In the Pennebaker and colleagues (2014) paper, the authors observed that students who performed academically better in college wrote their admissions essays in a more analytic and formal manner than those who performed academically worse. Linguistically, this meant students with high (versus low) grades in college used more articles and prepositions, but less storytelling words like auxiliary verbs, personal pronouns, and conjunctions. This finding helped to create an analytic thinking index, which is a bipolar continuum that approximates a person's communicative and thinking style (for a sample of studies using this index, see Abe, 2023; Chung & Pennebaker, 2018; Figueiredo & Devezas, 2021; Monzani et al., 2021; Seih & Lepicovsky, 2020). Together, style words and the analytic thinking index are critical to mark cognitive characteristics related to one's psychology and thus, they are essential for approximating the internal processing of the human mind.

Finally, the third class of dimensions investigated in most psychology of language studies is verbal structure, which considers how words and sentences are organized and arranged. Verbal structure is often measured through readability (Flesch, 1948) and lexical diversity metrics like type-token ratio (Richards, 1987; Templin, 1957) for a global assessment of a text's structure. In just one example, deception research by Markowitz and Hancock (2016) observed that papers written by scientists who engaged in data fraud were written in a more structurally complex and less readable manner than scientists who did not engage in data fraud. Therefore, psychological information about communicators can be obtained by not only observing what people communicate (content) and how they communicate (style), but also how the words and sentences within a text are generally arranged (structure).

**Self-Presentation, Warranting, and Language Patterns**

Recall, the primary interest of the current work is to examine how prolific sellers in an online marketplace communicate about their products — and by association, themselves — relative to those who are everyday sellers. It is therefore important to couch this work in theories of self-presentation to make the case that the text from a product description can reveal important psychological information about the seller. Self-presentation research in the social sciences is often attributed to early work by Goffman (1959), who argued that people often attempt to regulate how they appear to others (see also Snyder, 1974). Impression management can be achieved in many ways in the physical world, including but not limited to the amplification of physical traits (e.g., making the self appear taller, more attractive) in order to appear more likeable, interesting, and approachable to others (Goffman, 1967).

In digital spaces, online self-presentation strategies are equally critical. Much attention has been paid to self-presentation strategies in settings like online dating (Birnbaum et al., 2020; DeAndrea et al., 2012; Ellison et al., 2012; Markowitz & Hancock, 2018; Toma & Hancock, 2012), where language is a primary vehicle for moving to the next stage of a relationship (e.g., from profiles and messaging to a face-to-face meeting). Online daters who communicate more information, and information that is concrete to their prospects, tend to be rated as more trustworthy, for example (Toma & Hancock, 2012). Relatedly, more recent work has also observed how the self-descriptions on a physician's online profile relates to how they are perceived by patients (Markowitz, 2023b). Physicians who communicated in a more personalized and humanizing manner (e.g., using more self-references), while also using more indicators of verbal confidence (e.g., more certainty, less tentativeness), were rated higher, on average, than physicians who used less of those markers. In a follow-up experiment, physicians whose profiles contained more self-references and more verbal confidence were perceived as warmer and more competent (Fiske et al., 2002) than physicians whose profiles contained less "I"-words and verbal confidence. Together, the prior evidence suggests self-presentation is a process that people use to strategically manage the impressions they give off to others, and language patterns are essential for understanding how this dynamic occurs in online spaces.

A critical question in online self-presentation research is why some claims are believed while others are discounted. Warranting theory (Parks, 2011; Walther & Parks, 2002) provides insight into this puzzle, and broadly proposes receivers assign evidentiary value to information in proportion to how difficult it would be for a communicator to manipulate such information. Claims with information outside a person's control carry high warranting value (e.g., specs about a product), and therefore, claims generated entirely by the communicator or seller carry little warranting value. Experimental work supports this asymmetry, where self-presentations become more accurate when accountability is heightened and ground truth is available to receivers (DeAndrea et al., 2012). Further, the profile-as-promise framework from online dating research extends the logic by treating a self-presentation as a commitment that will be checked against offline behavior (Ellison et al., 2012), which can facilitate more honest or genuine disclosures.

Marketplace descriptions, like those represented on Reverb.com, provide a testing ground for this asymmetry. A claim about an instrument's specifications (e.g., its scale length, nut width, or pickup configuration) is anchored to an objective property of the instrument that the buyer can verify when receiving it. Any mismatch or inaccurate description can be communicated in seller feedback. A claim about how an instrument sounds, on the other hand, is subjective and not easily or readily falsifiable prior to purchase. Warranting theory therefore implies that objective and verifiable content should be more impactful than subjective content prior to purchase, and sellers who are prolific (e.g., they sell many products and are therefore verified often) should experience this dynamic more intensely than sellers who face it less often.

Together, self-presentation theory, warranting theory, and impression management research are especially important in the present setting for several reasons. In the online musical instrument marketplace under investigation, there is a bidirectional feedback mechanism. That is, buyers rate sellers and sellers rate buyers. Products that are incompletely or inappropriately described can put sellers at risk for decreased future sales. It is therefore crucial for sellers to accurately and fully characterize their product because such descriptions become a reflection not only of the item being sold, but also of the seller, their trustworthiness, and credibility (e.g., Rui, 2018). Sellers must put their best self forward in product descriptions to be perceived as a competent and reliable person on the marketplace. This bidirectional feedback mechanism, blended with the longitudinal nature of such evaluations (e.g., buyers can look at the full history of a seller's feedback), encourages descriptions that are rich, accurate, and reflective of the product and person (e.g., see Warkentin et al., 2010). The current paper uses Reverb.com as a key place to evaluate how seller status (i.e., prolific seller vs. everyday seller) is associated with verbal descriptions related to self-presentation, independent of the online seller's inventory size.

## The Current Paper

Against this backdrop, the current work uses a large corpus of product descriptions from

Reverb.com to evaluate how prolific sellers communicate relative to everyday sellers. Reverb is a specialized online marketplace for musical instruments and gear, and it offers several features that make it well-suited to the present aims. Sellers write open-ended descriptions of what they are offering, and because buyers cannot physically view, inspect, or play a product before purchase, these descriptions carry substantial evaluative weight. The text within each listing is also user-generated and unconstrained. Sellers decide what to emphasize, how to communicate, and how much of themselves to include alongside the object. Such descriptions are therefore product descriptions *and* self-presentation acts, offering a record of how people manage impressions when the stakes are high (e.g., trying to sell a product). Crucially, and perhaps most important for the aims of the current work, Reverb tracks selling volume for each person, which allows for a distinction to be made between prolific sellers from everyday sellers. Reverb is also a setting where verbal markers of self-presentation can be observed at scale.

Consistent with the *words-as-attention* framework that this study is derived from (Boyd & Markowitz, 2025; Boyd & Schwartz, 2021), the current work evaluates prolific and everyday seller language across content (what sellers communicate about), style (how they communicate), and verbal structure dimensions (how sellers' words are organized). Logic from warranting theory (Parks, 2011; Walther & Parks, 2002) offers a prediction for the first of these dimensions. Because prolific sellers describe products more often, and because descriptions are checked against the product upon delivery and recorded in public feedback, prolific sellers encounter verification more frequently than everyday sellers. Content that is anchored to the objective characteristics of the instrument can survive such scrutiny, whereas descriptions that rely on subjective impressions cannot be evaluated by a buyer before purchasing. Thus, objective claims and descriptions carry more warranting value than subjective claims and descriptions. Prolific sellers should therefore concentrate on the former rather than the latter. Accordingly, the following hypothesis and research questions are posed:

$H_1$: Prolific sellers will attend more to verifiable, objective product attributes and less to unverifiable, subjective attributes than everyday sellers.

Note, warranting theory does not make a direct prediction for communicative style nor verbal structure. Prior self-presentation research suggests impression management in digital settings is achieved through a range of linguistic markers (e.g., Markowitz, 2023b; Toma & Hancock, 2012) without specifying their direction for this specific setting or marketplace. These dimensions are therefore examined in an exploratory manner:

$RQ_{1a\text{-}b}$: How do the (a) style and (b) structural patterns in product descriptions differ between prolific and everyday sellers?

To address the prior prediction and research question, 1.6 million Reverb listings were analyzed for their verbal content, style, and structural patterns, while accounting for other metadata-related features like visual elements (e.g., photos, videos) and seller characteristics (e.g., if Reverb designates a seller as "preferred"). This helped to isolate the impact of language alone in the process of managing impressions and self-presenting. Content patterns were measured using a bottom-up approach for thematic extraction (Chung & Pennebaker, 2008; Markowitz, 2021), communicative style was approximated by an analytic thinking index used in prior work (Pennebaker et al., 2014), and verbal structure was measured with readability and type-token ratio metrics (Flesch, 1948; Richards, 1987).

# Method

## Data Collection and Preprocessing

An initial collection of 2.1 million product listings was obtained from Reverb.com, a peer-to-peer musical instrument and equipment marketplace. Using Reverb's dedicated API, several pieces of data were extracted from all available product listings: (1) the product description in text form, (2) the number of days that elapsed since the product was listed, (3) the price of the product (in US dollars), (4) the primary category of the listing, of which there were fourteen, (5) visual metrics like photo count and video count, (6) the state of the listing (i.e., live, ended, sold), and (7) engagement metrics, including the number of offers on the product (e.g., a count representing how many buyers made an offer to purchase the product), the number of page views (e.g., a count representing the number of times the product page was visited), and the number of "saves" (e.g., a count representing the number of times a user wanted remember this product for easy access upon returning to Reverb).

Only live listings were retained because ended and sold listings are infrequently available and often archived from the API. Product descriptions were preprocessed by automatically removing non-English texts (e.g., Ooms, 2020). This resulted in a final sample of 1,628,876 Reverb listings that covered over 300 million words (*M* = 201.97 words per listing, *SD* = 220.02 words per listing, *Mdn* = 137 words per listing, min = 1 word, max = 11,460 words).

**Automated Text Analysis**

All product descriptions were analyzed with Linguistic Inquiry and Word Count (LIWC) software (Pennebaker et al., 2022). LIWC is a computer program that counts words as a percentage of the total word count per text. It contains an internal dictionary of human-validated categories that reflect social (e.g., words related to family), psychological (e.g., words related to emotion, cognition), and part of speech dimensions (e.g., articles, pronouns) reflecting the human condition. For example, the following sentence — "I have a custom guitar ready for sale." — contains 8 words and counts the following LIWC categories, including but not limited to self-references (i.e., *I*, or 12.5% of the total word count), articles (i.e., *a*, or 12.5% of the total word count), and money-related terms (i.e., *sale*, or 12.5% of the total word count).

***Content Patterns: The Meaning Extraction Method***

Content patterns on Reverb were assessed using a bottom-up topic modeling approach called the Meaning Extraction Method (Chung & Pennebaker, 2008; Markowitz, 2021). This approach begins by removing function words (e.g., articles, prepositions) and low base-rate words from a text. Next, content words like nouns and verbs are retained and each text assigned a binary score (1 = a content word is present, 0 = a content word is absent). Unigrams (single words), bigrams (two-word phrases), and trigrams (three-word phrases) were selected as the textual units in the meaning extraction process. Then, using Principal Component Analysis with varimax rotation, content words clustered, statistically, to retain coherent themes that emerged from the data. Themes were retained according to best practices (Chung & Pennebaker, 2008; Foxman et al., 2021; Markowitz, 2021), including the idea that each incremental theme should contribute a meaningful amount of variance explained, associated eigenvalues should be greater than one, and each theme should be face-validly interpretable. A total of 8 themes were retained during this thematic extraction process. Themes were saved as standardized regression weights ($M = 0$, $SD = 1$) for future inclusion in statistical models.

***Style Patterns: Analytic Thinking***

LIWC scores on the analytic thinking index were produced across all Reverb product descriptions. The analytic thinking index is a standardized composite of 8 function word categories, including positive loadings for articles and prepositions, and negative loadings for auxiliary verbs, adverbs, personal pronouns, impersonal pronouns, negations, and conjunctions. Raw scores are converted to a standardized metric upon comparing such values to comparison corpora (Boyd et al., 2022). Low scores reflect a dynamic, storytelling, and informal communication style; high scores reflect a hierarchical and formal communication style. Scores around 50 are "middle of the road."

#### ***Structural Patterns: Readability and Type-Token Ratio***

Two dimensions were used to evaluate the structural properties of Reverb listings. The first dimension is Flesch Reading Ease (Flesch, 1948), which is a standard readability metric that accounts for the number of words per sentence and syllables per word in a piece of text. High scores on this dimension indicate a more readable text (e.g., fewer words per sentence, fewer syllables per word) compared to low scores on this dimension. The second structural metric is type-token ratio (TTR), which is a measure of lexical diversity. TTR assesses the degree to which a text contains a high proportion of unique and varied terms (high TTR) or a high proportion of consistency or reuse of terms (low TTR). Both readability and TTR were measured with the *quanteda.textstats* package in R (Benoit et al., 2021).

### **Prolific Seller Identification**

To identify how the communicative style of prolific sellers on Reverb compared to everyday sellers, the total number of reviews associated with each seller were obtained (e.g., formally called *feedback* on Reverb), and analyses were restricted to the 14,827 sellers with at least five active listings to avoid one-off transactions. This created a dichotomous independent variable called seller type in which a prolific seller was defined as falling in the top decile (10%) of feedback (this corresponds to 697 rated transactions), which is an approach broadly consistent with prior work (Mierlo, 2014; Simoni et al., 2020; Wojcik & Hughes, 2019). This classified 1,484 sellers as *prolific sellers* who accounted for 1,131,989 listings (69.5% of the sample). The remaining 13,343 sellers, called *everyday sellers*, accounted for 496,887 listings.

**Analytic Plan and Covariates**

Using the *fixest* package in R (Berge et al., 2026), dependent variables were language patterns reflecting content, style, or structural characteristics. The focal predictor had two levels: prolific seller vs. everyday seller (reference group), and standard errors were clustered at the seller level. Several metadata covariates were included to isolate the association between seller status (i.e., prolific seller vs. everyday seller) and language patterns. These metadata covariates were: (1) product price, (2) inventory size of the seller, (3) photo and video count of the listing, (4) active days of the listing, (5) primary category of the listing, (6) the condition of the listing, (7) preferred seller status, and (8) listing engagements (i.e., views, offers, saves).[1]

**Price.** Price of the product was logged, $ln(X + 1)$, and standardized within primary product category. Expensive items may warrant more detailed, specification-heavy descriptions regardless of who is selling them. Standardizing within category is necessary because absolute price is inherently different across product categories (e.g., the ceiling price of a microphone is different than the ceiling price of vintage guitar).

**Inventory size.** The number of active listings for each seller was logged, $ln(X + 1)$. Inventory size proxies the scale of a seller's operation, as scale may shape how sellers write. A seller managing hundreds of listings may have little time to compose each one and is more likely to lean on templates, boilerplate, and copy-and-paste descriptions. A seller with a handful of items can tailor language to each product and spend more time on them. Controlling for inventory size separates the effects of interest from the stylistic signature of selling at volume.

**Photo and video count.** Listings with more visual evidence associated with their products may need to use words less or describe their product in different levels of detail. Including variables that account for the visual nature of marketplace listings ensures that any effect of communication style is not a byproduct of how much visual content is in the listing.

[1] This modeling approach was selected, compared to a linear mixed model, because unobserved seller-level characteristics are plausibly correlated with feedback volume, which would violate the random-effects assumption. Plus, intraclass correlation coefficients across all key dependent variables in this study were between 0.13 and 0.63, supporting the decision to cluster standard errors at the seller level.

**Active days of listing.** Recall, the present Reverb corpus consists of live listings and therefore, the number of active days is a function of publication date. Including fixed effects for active days ensures comparisons are made among listings with the same listing age.

**Primary category and condition.** There are fourteen primary categories for Reverb products (i.e., electric guitars, *n* = 170,488; accessories, *n* = 264,678; acoustic guitars, *n* = 74,803; amps, *n* = 46,785; band and orchestra, *n* = 62,612; bass guitars, *n* = 37,354; DJ and lighting gear, *n* = 26,763; drums and percussion, *n* = 286,754; effects and pedals, *n* = 159,955; folk instruments, *n* = 31,887; home audio, *n* = 32,960; keyboards and synths, *n* = 66,047; parts, *n* = 218,315; pro audio, *n* = 149,475). Controlling for primary category accounts for emergent between-product differences (e.g., a microphone will likely be described differently than a guitar). The product condition variable contained nine categories (i.e., B-stock, *n* = 20,483; brand new, *n* = 1,192,887; excellent, *n* = 123,480; fair, *n* = 11,182; good, *n* = 69,136; mint, *n* = 67,770; non-functioning, *n* = 5,973; poor, *n* = 1,690; very good, n = 136,275) also accounted for heterogeneity in descriptions related to the shape of the product.

**Preferred seller status.** Reverb has a badge associated with sellers who “have established a solid track record” (Reverb, 2026). Accounting for this binary variable isolated what accumulated feedback adds beyond the badge for prolific versus everyday sellers.

**Engagements**. The number of engagements associated with each listing was included to account for differences in marketplace attention. Listings that receive more engagements (i.e., more views, saves, or offers) may differ systematically from listings with less engagements in ways that are also associated with how sellers describe their products (for a related example using simple language, see Markowitz & Shulman, 2021). To create an engagement index, the three engagement metrics were log-transformed, *ln*(X + 1) and then separately standardized (*z*-scored). The engagement index was then calculated as the mean of the three standardized scores for each listing. Intercorrelations between the log-transformed components were positive and statistically significant (.299 < *r*s < .725).

## Results

Descriptive statistics for all variables and their intercorrelations are in Table 1.[2]

### Content Patterns

The evidence in Table 2 describes the eight themes that were extracted. These eight themes were further grouped into three higher-order collections:

(1) *Product specifications*: guitar body (Component 1), drums (Component 3), electronics (Component 5), response latency (Component 7), and power supply information (Component 8)

(2) *Post-purchase information*: shipping (Component 2) and customer service (Component 4)

(3) *Acoustics*: tone (Component 6)

The results in Table 3 describe the themes most systematically related to seller type. There are two findings that were statistically significant at the 5% level. First, compared to everyday sellers, prolific sellers focus more on specific details of guitar body like *frets*, *scale*

[2] Because the 90th-percentile cut-point was an exploratory threshold, every model was re-estimated with the prolific vs. everyday threshold set at the 50th and 75th percentiles of the seller-level feedback distribution (holding the sample, covariates, fixed effects, and clustering constant). Coefficient signs were consistent across all three cut-points, and the 95% confidence intervals overlapped for every outcome. Therefore, the reported associations do not depend on where the threshold is drawn. These consistencies indicate the robustness of the pattern rather than a common effect size.

*length*, and the guitar *bridge* ($p$ = .001). It is critical to observe that among the top words representing this component, none of them related to the look or feel of the instrument. Instead, prolific sellers focused on descriptions of functional, structural, or playability-related features instead of aesthetics. The second theme that was significantly related to seller type was tone ($p$ = .039), where prolific sellers focused less on tone-related terms like *rich*, *warm*, *smooth*, and *clarity* compared to everyday sellers. In other words, prolific sellers focus less on sound-related subjective claims about an instrument. Thus, $H_1$ was partially supported.

**Style Patterns**

Addressing $RQ_{1a}$, the relationship between seller type and analytic thinking was negative and statistically significant ($B$ = -1.52, $SE$ = 0.65, $t$ = -2.35, $p$ = .019, Cohen's $f^2$ = 0.001). This pattern suggests prolific sellers have a more informal, dynamic, and narrative-like communication style relative to everyday sellers.

Recall, the analytic thinking measure is an eight-category composite and therefore, it is important to identify the dimension(s) driving this significant negative effect. Eight fixed effect models were run, correcting $p$-values by multiplying them by the number of tests run to avoid Type I errors, and one significant relationship was observed. The relationship between seller type and the personal pronouns was significant after multiple comparisons corrections ($B$ = 0.38, $SE$ = 0.13, $t$ = 3.01, $p$ = .003; corrected to $p$ = .021). Therefore, prolific sellers personalize and humanize their listings by using more terms like *I*, *you*, and *she* than everyday sellers.

**Structural Patterns**

Addressing $RQ_{1b}$, the relationship between seller type and readability was not statistically significant ($B$ = 0.49, $SE$ = 0.81, $t$ = 0.60, $p$ = .549, Cohen's $f^2$ < 0.001). The relationship between seller type and type-token ratio was statistically significant ($B$ = -0.02, $SE$ = 0.01, $t$ = - 2.13, $p$ = .033, Cohen's $f^2$ = 0.002). It is important to note, however, that controlling for the number of tokens in the listing led to a non-significant result between seller type and TTR ($B$ = 0.003, $SE$ = 0.003, $t$ = 1.00, $p$ = .317). TTR was also negatively correlated with log token count ($r$ = -0.90, $p$ < .001). Using log token count as the dependent variable in the prior modeling technique led to a positive relationship between prolific seller status and token count ($B$ = 0.15, $SE$ = 0.05, $t$ = 2.75, $p$ = .006), or the result that there is approximately 16% more text in prolific versus everyday seller listings. Therefore, the structural difference between prolific and everyday sellers may reflect description length rather than lexical diversity, and prolific seller status is a positive predictor of description length.

## Discussion

The current work used 1.6 million product listings from an online music product marketplace to evaluate how different seller types (i.e., prolific sellers vs. everyday sellers) communicate about their products and reveal important psychological aspects of their self-presentation. The results, evaluated across content, style, and structural dimensions of language use, revealed what sellers psychologically attend to in their descriptions. Eight coherent themes emerged in the meaning extraction analysis across three higher-order collections: product specifications, post-purchasing information, and acoustic information. Prolific sellers tended to focus less on acoustic-related parts of their descriptions relative to everyday sellers; instead, they often focused on functional and structural aspects of instruments that seasoned players would typically care about (perhaps as a representation of their expertise). Stylistically, prolific sellers wrote in a more narrative-like manner than everyday sellers. This pattern suggests that *what* prolific sellers focus on tends to be quite technical, but *how* they communicate this information is like a story and informal (and perhaps more approachable) to potential buyers. Finally, the structural features of prolific sellers were different from everyday sellers at least for the diversity of language dimensions represented in the product descriptions. Prolific sellers wrote longer descriptions, and once description length was accounted for, differences in lexical diversity (via TTR) were not obtained.

There are several important contributions to self-presentation theory and impression management research based on the present work. First, the results suggest impression management in a marketplace setting operates across distinct communicative behaviors. Prolific sellers and everyday sellers differed in what and how they focused their attention (e.g., content and style patterns), and how they arranged their descriptions (e.g., structural patterns), but these two sets of results moved in opposite directions with respect to formality. Content patterns among prolific sellers were technical and specs-oriented for the different musical products (e.g., emphasizing frets, scale length, and bridge hardware) instead of the look, feel, or sound of an instrument. Style patterns, on the other hand, were less analytic and more narrative-like with a particularly elevated rate of personal pronouns. Much self-presentation scholarship treats impression management as a single performance that is either more or less formal, or more or less personalized (Goffman, 1959; Snyder, 1974). The present evidence suggests that communicators can be technical *and* personal at the same time, which may serve as signals of competence that associate with their prolific selling status (though future experimental work would be helpful to validate this assertion). This technical and personal distinction is consistent with prior work that observed how self-references and verbal confidence jointly predict perceptions of warmth and competence for physicians (Markowitz, 2023b). These two dimensions of social- and person-perception are indeed carried by distinct features of the same collection of verbal behavior.

Second, it is also important to dissect the content effects further and opine about their underlying relationship to seller status. The content effects point to accountability as a possible organizing principle behind how prolific sellers self-present. Prolific sellers attended less to tone-related terms and more to playability-related features. The distinction between these two types of claims may, according to prior work, serve as warrants in an online environment (e.g., signals in an online environment used to assess the accuracy of communicated information; Parks, 2011; Walther & Parks, 2002). Acoustic descriptors are subjective and largely unfalsifiable online, whereas claims related to specs are verifiable against the object once the buyer receives it. Therefore, in this sense, content about the specs of an instrument or product carries more warranting value than acoustic content because specs are anchored to a property of the object that a buyer will encounter directly (and the seller cannot control how the claim is evaluated). However, subjective claims about how an instrument sounds are difficult to disconfirm online, and it is therefore easy for any seller to describe or feign. Warranting theory suggests buyers should discount the subjective claims and prize the objective claims, which is consistent with prolific sellers having converged on this asymmetry.

Related to self-presentation, the content findings can also be evaluated through the profile-as-promise perspective from online dating research (Ellison et al., 2012). A product's description on a marketplace is a commitment that it will eventually be checked against reality and ground truth, and sellers who are prolific face that check more often than everyday sellers. Therefore, a seller who writes about an instrument's acoustics and tonality cannot easily be contradicted online; more objective descriptions about specs can be contradicted upon receiving the items and such inconsistencies can be provided in seller feedback. The observed patterns are therefore consistent with the idea that repeated exposure to verification (e.g., prolific sellers are evaluated and rated more than everyday sellers) pushes self-presentation toward claims that can survive such pressure and scrutiny. This extends self-presentation theory, which is typically characterized by  one-shot encounters like in the online dating literature (Birnbaum et al., 2020; Ellison et al., 2012; Toma & Hancock, 2012), toward settings where the same self is presented hundreds or thousands of times to an accumulating audience.

Third, prolific sellers wrote longer descriptions than everyday sellers (e.g., roughly 16% more text after accounting for covariates). Classic impression management research assumes a performance is tailored to a particular audience in a particular moment to achieve various self-presentation goals (Goffman, 1959, 1967). However, the present data suggest prolific sellers do not rest on their reputation as volume sellers. Instead, prolific sellers have longer descriptions than everyday sellers. This length finding complements the content effects. That is, longer descriptions afford more room for the verifiable, specs-oriented claims that prolific sellers communicated more than everyday sellers, which is what warranting logic would predict of sellers who face verification from various purchasers on a repeated basis. It is unclear whether self-presentation research has properly contended with impression management when it is reproduced at scale like on Reverb or similar marketplaces. More work is required to identify whether these patterns hold across other online marketplaces and self-presentation settings.

**Practical Implications**

There are several practical implications worth highlighting about this work. For sellers, the results provide empirical evidence to articulate what and how prolific sellers communicate compared to everyday sellers. The present study cannot establish causality, and the observed differences were small in magnitude. However, the data provide practical evidence regarding how more experienced sellers write compared to less experienced sellers. For platforms, the results suggest prolific sellers lean toward verifiable content and specs over speculation. If specs and objective details are markers of expertise, any platform-supplied advice may encourage sellers to communicate in a way that claims can be verified and objectively assessed prior to purchase. The tone result also suggests subjective claims may be characteristics that novice sellers turn to when they have few concrete points to make about a product.

**Limitations and Future Directions**

There are several limitations of this work that deserve greater attention in future research. First, the effect sizes observed here were small in magnitude. Second, prolific seller status was operationalized through accumulated feedback rather than a direct measure of sales volume. Feedback accrues with both transaction volume and platform tenure, though the two cannot be separated in the present data. Future work with transaction-level data, if possible to obtain, could disentangle experience from actual sales output, which may relate to language in different ways.

Third, the study is observational, the claims are not causal, and the design excluded items that already sold since Reverb does not index these listings systematically for analysis via their API. The present work therefore describes how seller types differ in their self-presentation via language patterns, but it cannot speak to whether these differences translate into sales. Finally, the corpus was restricted to English descriptions from a single specialized marketplace. Reverb's buyers likely have domain-level knowledge prior to purchasing a product, which may make specs and related details more valuable there than in general-purpose marketplaces. Whether the content, style, and structural patterns observed here generalize to marketplaces with less expert buyers is an open question that deserves additional treatment and future study.

**Table 1**

*Intercorrelations Between Variables*

| Variable | *M* | *SD* | 1 | 2 | 3 | 4 | 5 | 6 | 7 | 8 | 9 | 10 |
|---|---|---|---|---|---|---|---|---|---|---|---|---|
| 1. C1 | 0.00 | 1.00 | | | | | | | | | | |
| 2. C2 | 0.00 | 1.00 | -.00<br>[-.00, .00] | | | | | | | | | |
| 3. C3 | 0.00 | 1.00 | -.00<br>[-.00, .00] | -.00<br>[-.00, -.00] | | | | | | | | |
| 4. C4 | 0.00 | 1.00 | -.00<br>[-.00, .00] | -.00<br>[-.00, .00] | -.00<br>[-.00, .00] | | | | | | | |
| 5. C5 | 0.00 | 1.00 | -.00<br>[-.00, .00] | -.00<br>[-.00, .00] | -.00<br>[-.00, .00] | -.00<br>[-.00, .00] | | | | | | |
| 6. C6 | 0.00 | 1.00 | .00<br>[-.00, .00] | -.00<br>[-.00, .00] | -.00<br>[-.00, .00] | -.00<br>[-.00, .00] | -.00<br>[-.00, .00] | | | | | |
| 7. C7 | 0.00 | 1.00 | -.00<br>[-.00, .00] | -.00<br>[-.00, .00] | -.00<br>[-.00, .00] | -.00<br>[-.00, .00] | -.00<br>[-.00, .00] | -.00<br>[-.00, .00] | | | | |
| 8. C8 | 0.00 | 1.00 | -.00<br>[-.00, .00] | -.00<br>[-.00, .00] | -.00<br>[-.00, .00] | -.00<br>[-.00, .00] | -.00<br>[-.00, .00] | -.00<br>[-.00, .00] | -.00<br>[-.00, .00] | | | |
| 9. Analytic thinking | 86.59 | 14.21 | .10**<br>[.10, .10] | -.19**<br>[-.20, -.19] | .09**<br>[.09, .09] | -.19**<br>[-.19, -.19] | .12**<br>[.12, .12] | .08**<br>[.08, .09] | .06**<br>[.06, .06] | .02**<br>[.02, .02] | | |
| 10. Readability | 46.33 | 25.51 | -.20**<br>[-.20, -.20] | .04**<br>[.04, .04] | -.00**<br>[-.00, -.00] | .08**<br>[.08, .09] | -.13**<br>[-.14, -.13] | -.04**<br>[-.04, -.04] | -.10**<br>[-.10, -.10] | -.05**<br>[-.05, -.05] | -.25**<br>[-.25, -.25] | |
| 11. TTR | 0.71 | 0.15 | -.19**<br>[-.19, -.18] | -.36**<br>[-.36, -.36] | -.27**<br>[-.27, -.27] | -.15**<br>[-.15, -.15] | -.22**<br>[-.23, -.22] | -.13**<br>[-.13, -.12] | -.12**<br>[-.12, -.12] | -.08**<br>[-.08, -.08] | -.08**<br>[-.08, -.07] | .16**<br>[.16, .16] |

*Note*. C1-C8 = Component 1 through Component 8, which are detailed in Table 2. Intercorrelations between C1-C8 are zero because they are fully orthogonal (varimax rotation). Numbers in brackets are 95% Confidence Intervals.

**Table 2**

*Content Patterns from the Meaning Extraction Method*

| C1 | | C2 | | C3 | | C4 | |
|---|---|---|---|---|---|---|---|
| Guitar body | | Shipping | | Drums | | Customer service | |
| λ | % | λ | % | λ | % | λ | % |
| 18.15 | 6.67 | 9.34 | 3.43 | 5.84 | 2.15 | 4.95 | 1.82 |
| Word | Loading | Word | Loading | Word | Loading | Word | Loading |
| nut | 0.834 | shipped | 0.550 | cymbals | 0.721 | items | 0.663 |
| frets | 0.830 | cost | 0.549 | cymbal | 0.691 | contact | 0.621 |
| scale | 0.815 | listing | 0.546 | hand | 0.642 | item | 0.557 |
| neck | 0.814 | sure | 0.533 | crafted | 0.642 | $ | 0.546 |
| scale length | 0.786 | materials | 0.530 | details | 0.628 | note | 0.517 |
| fingerboard | 0.781 | due | 0.528 | brand | 0.565 | shipping | 0.465 |
| body | 0.767 | send | 0.521 | years | 0.554 | purchase | 0.425 |
| radius | 0.765 | lower | 0.519 | 22 | 0.456 | country | 0.425 |
| bridge | 0.753 | find | 0.515 | available | 0.452 | sold | 0.423 |
| nut width | 0.735 | parts | 0.505 | traditional | 0.396 | questions | 0.407 |

| C5 | | C6 | | C7 | | C8 | |
|---|---|---|---|---|---|---|---|
| Electronics | | Tone | | Response latency | | Power | |
| λ | % | λ | % | λ | % | λ | % |
| 3.71 | 1.36 | 2.95 | 1.08 | 2.61 | 0.96 | 2.61 | 0.96 |
| Word | Loading | Word | Loading | Word | Loading | Word | Loading |
| output | 0.652 | rich | 0.519 | frequency | 0.683 | power supply | 0.927 |
| input | 0.651 | warm | 0.515 | response | 0.619 | supply | 0.921 |
| jack | 0.605 | sustain | 0.451 | dynamic | 0.448 | power | 0.593 |
| signal | 0.457 | smooth | 0.373 | noise | 0.385 | | |
| level | 0.453 | tonal | 0.340 | low | 0.326 | | |
| switch | 0.351 | tones | 0.332 | | | | |
| audio | 0.325 | delivers | 0.313 | | | | |
| effects | 0.274 | clarity | 0.307 | | | | |
| dual | 0.271 | | | | | | |

*Note*. λ = Eigenvalues. % = percent variance explained by each component. C1-C8 = Component 1 through Component 8. Only the top 10 words within each component are represented here for readability purposes.

**Table 3**

*Content Results from the Fixed Effect Models*

| Theme | Description | Predictor | *B* | *SE* | *t* | *p* | Cohen's $f^2$ | 95% CI | $R^2_{within}$ |
|---|---|---|---|---|---|---|---|---|---|
| 1 | Guitar body | Seller type: Prolific | 0.068 | 0.021 | 3.23 | .001 | 1.31E-03 | [0.027, 0.109] | 0.0189 |
| 1 | Guitar body | Price | 0.059 | 0.007 | 8.98 | < .001 | 8.31E-03 | [0.046, 0.072] | 0.0189 |
| 1 | Guitar body | Inventory size | 0.006 | 0.009 | 0.67 | .506 | 1.70E-04 | [-0.011, 0.023] | 0.0189 |
| 1 | Guitar body | Photo count | 0.010 | 0.003 | 3.39 | < .001 | 4.03E-03 | [0.004, 0.016] | 0.0189 |
| 1 | Guitar body | Video count | 0.002 | 0.018 | 0.11 | .912 | 0.00E+00 | [-0.034, 0.038] | 0.0189 |
| 1 | Guitar body | Engagement index | 0.003 | 0.009 | 0.3 | .763 | 1.00E-05 | [-0.015, 0.020] | 0.0189 |
| 2 | Shipping | Seller type: Prolific | 0.189 | 0.097 | 1.95 | .051 | 5.53E-03 | [-0.001, 0.378] | 0.0518 |
| 2 | Shipping | Price | -0.003 | 0.021 | -0.13 | .894 | 1.00E-05 | [-0.044, 0.038] | 0.0518 |
| 2 | Shipping | Inventory size | 0.084 | 0.051 | 1.64 | .100 | 1.90E-02 | [-0.016, 0.183] | 0.0518 |
| 2 | Shipping | Photo count | 0.020 | 0.002 | 8.43 | < .001 | 8.46E-03 | [0.015, 0.024] | 0.0518 |
| 2 | Shipping | Video count | 0.037 | 0.032 | 1.15 | .249 | 1.80E-04 | [-0.026, 0.099] | 0.0518 |
| 2 | Shipping | Engagement index | -0.006 | 0.011 | -0.56 | .577 | 2.00E-05 | [-0.028, 0.015] | 0.0518 |
| 3 | Drums | Seller type: Prolific | -0.075 | 0.047 | -1.59 | .112 | 1.22E-03 | [-0.168, 0.018] | 0.0503 |
| 3 | Drums | Price | 0.095 | 0.02 | 4.67 | < .001 | 1.61E-02 | [0.055, 0.135] | 0.0503 |
| 3 | Drums | Inventory size | 0.077 | 0.033 | 2.37 | .018 | 2.27E-02 | [0.013, 0.142] | 0.0503 |
| 3 | Drums | Photo count | 0.003 | 0.004 | 0.96 | .335 | 3.50E-04 | [-0.003, 0.010] | 0.0503 |
| 3 | Drums | Video count | 0.150 | 0.042 | 3.6 | < .001 | 4.13E-03 | [0.068, 0.231] | 0.0503 |
| 3 | Drums | Engagement index | -0.004 | 0.008 | -0.52 | .602 | 1.00E-05 | [-0.020, 0.012] | 0.0503 |
| 4 | Customer service | Seller type: Prolific | 0.010 | 0.083 | 0.12 | .904 | 1.00E-05 | [-0.152, 0.172] | 0.0113 |
| 4 | Customer service | Price | 0.003 | 0.017 | 0.2 | .841 | 1.00E-05 | [-0.030, 0.037] | 0.0113 |
| 4 | Customer service | Inventory size | 0.036 | 0.024 | 1.51 | .131 | 2.61E-03 | [-0.011, 0.082] | 0.0113 |
| 4 | Customer service | Photo count | 0.021 | 0.003 | 5.99 | < .001 | 7.16E-03 | [0.014, 0.027] | 0.0113 |
| 4 | Customer service | Video count | 0.049 | 0.042 | 1.17 | .242 | 2.40E-04 | [-0.033, 0.131] | 0.0113 |
| 4 | Customer service | Engagement index | -0.010 | 0.021 | -0.47 | .635 | 4.00E-05 | [-0.050, 0.031] | 0.0113 |
| 5 | Electronics | Seller type: Prolific | 0.011 | 0.026 | 0.41 | .679 | 1.00E-05 | [-0.040, 0.062] | 0.0049 |
| 5 | Electronics | Price | 0.053 | 0.009 | 6.07 | < .001 | 2.71E-03 | [0.036, 0.070] | 0.0049 |
| 5 | Electronics | Inventory size | -0.004 | 0.01 | -0.37 | .714 | 3.00E-05 | [-0.023, 0.016] | 0.0049 |
| 5 | Electronics | Photo count | 0.002 | 0.002 | 1.16 | .246 | 1.00E-04 | [-0.002, 0.006] | 0.0049 |
| 5 | Electronics | Video count | 0.011 | 0.026 | 0.44 | .661 | 1.00E-05 | [-0.040, 0.063] | 0.0049 |
| 5 | Electronics | Engagement index | 0.037 | 0.014 | 2.56 | .011 | 5.30E-04 | [0.009, 0.065] | 0.0049 |
| 6 | Tone | Seller type: Prolific | -0.057 | 0.028 | -2.06 | .039 | 3.50E-04 | [-0.112, -0.003] | 0.0026 |
| 6 | Tone | Price | 0.020 | 0.01 | 1.95 | .051 | 3.40E-04 | [-0.000, 0.040] | 0.0026 |
| 6 | Tone | Inventory size | 0.021 | 0.008 | 2.5 | .012 | 7.90E-04 | [0.004, 0.037] | 0.0026 |
| 6 | Tone | Photo count | -0.004 | 0.003 | -1.46 | .144 | 2.00E-04 | [-0.009, 0.001] | 0.0026 |
| 6 | Tone | Video count | 0.105 | 0.03 | 3.48 | < .001 | 1.00E-03 | [0.046, 0.164] | 0.0026 |
| 6 | Tone | Engagement index | -0.008 | 0.009 | -0.91 | .365 | 3.00E-05 | [-0.027, 0.010] | 0.0026 |
| 7 | Response latency | Seller type: Prolific | 0.000 | 0.022 | 0.01 | .993 | 0.00E+00 | [-0.043, 0.043] | 0.0119 |
| 7 | Response latency | Price | 0.099 | 0.014 | 7.19 | < .001 | 9.33E-03 | [0.072, 0.126] | 0.0119 |
| 7 | Response latency | Inventory size | 0.003 | 0.009 | 0.32 | .751 | 2.00E-05 | [-0.015, 0.021] | 0.0119 |
| 7 | Response latency | Photo count | 0.000 | 0.003 | 0.07 | .941 | 0.00E+00 | [-0.006, 0.007] | 0.0119 |
| 7 | Response latency | Video count | 0.103 | 0.035 | 2.93 | .003 | 1.05E-03 | [0.034, 0.172] | 0.0119 |
| 7 | Response latency | Engagement index | -0.019 | 0.014 | -1.33 | .184 | 1.40E-04 | [-0.048, 0.009] | 0.0119 |
| 8 | Power | Seller type: Prolific | -0.039 | 0.021 | -1.87 | .061 | 1.80E-04 | [-0.080, 0.002] | 0.003 |
| 8 | Power | Price | 0.018 | 0.011 | 1.58 | .114 | 3.00E-04 | [-0.004, 0.039] | 0.003 |
| 8 | Power | Inventory size | 0.022 | 0.014 | 1.54 | .123 | 9.60E-04 | [-0.006, 0.049] | 0.003 |
| 8 | Power | Photo count | 0.007 | 0.002 | 3.52 | < .001 | 9.20E-04 | [0.003, 0.012] | 0.003 |
| 8 | Power | Video count | -0.011 | 0.018 | -0.6 | .546 | 1.00E-05 | [-0.046, 0.024] | 0.003 |
| 8 | Power | Engagement index | 0.028 | 0.014 | 1.91 | .056 | 3.00E-04 | [-0.001, 0.056] | 0.003 |

*Note*. Models include fixed effects for listing days, primary category, condition, and preferred-seller status; these are absorbed and not reported.